\pdfoutput=1

\documentclass[11pt]{article}

\usepackage[final]{acl}

\usepackage{times}
\usepackage{latexsym}

\usepackage[T1]{fontenc}

\usepackage[utf8]{inputenc}

\usepackage{microtype}

\usepackage{inconsolata}

\usepackage{graphicx}

\usepackage{amsmath}
\usepackage[ruled,vlined]{algorithm2e}
\usepackage{algorithm2e}
\usepackage{float}
\usepackage{mdframed}
\usepackage{geometry} \usepackage{array, makecell}%
\setcellgapes{3pt}
\usepackage{bbm}

\title{PromptMind Team at MEDIQA-CORR 2024: Improving Clinical Text Correction with Error Categorization and LLM Ensembles}

\author{Satya K Gundabathula \\
  \texttt{satyakesav123@gmail.com} \\\And
  Sriram R Kolar \\
  \texttt{sriramrakshithkolar@gmail.com} \\}

\begin{document}
\maketitle
\begin{abstract}
This paper describes our approach to the MEDIQA-CORR shared task, which involves error detection and correction in clinical notes curated by medical professionals. This task involves handling three subtasks: detecting the presence of errors, identifying the specific sentence containing the error, and correcting it. Through our work, we aim to assess the capabilities of Large Language Models (LLMs) trained on a vast corpora of internet data that contain both factual and unreliable information. We propose to comprehensively address all subtasks together, and suggest employing a unique prompt-based in-context learning strategy. We will evaluate its efficacy in this specialized task demanding a combination of general reasoning and medical knowledge. In medical systems where prediction errors can have grave consequences, we propose leveraging self-consistency and ensemble methods to enhance error correction and error detection performance.
\end{abstract}

\section{Introduction}

With rapid advancements in Natural Language Processing (NLP), we are witnessing a surge of its applications across various domains, including healthcare. Incorporating NLP in clinical settings brings about a multitude of advantages. It enhances clinical decision-making through advanced support, making health information more accessible,  streamlines documentation, and accelerates research initiatives \citep{nlpbenefits}. These developments contribute to improved patient care, reduced healthcare costs, and alleviated physician burnout.

NLP for healthcare applications pose inherent challenges due to the need for medical expertise. However, advancements in LLMs trained on internet data including medical information have significantly enhanced their knowledge and reasoning capabilities, enabling them to tackle more complex problems in the healthcare domain involving text processing and generation. Few recent applications in healthcare include information extraction, question answering, summarization, and translation, all while comprehending intricate medical knowledge \citep{healthcarellms}. Despite these advancements, safety and accuracy remain major concerns as training data may contain unreliable and misleading information that could have adverse effects if not handled appropriately. Nevertheless, the effective utilization of these LLMs has the potential to revolutionize healthcare and bring immense benefits to society \citep{landscape}.

In the healthcare industry, there is a need for automated systems capable of efficiently analyzing and interpreting clinical texts that improves patient’s safety, quality of care and costs.  Processing the clinical texts presents a unique and significant challenge due to the complexities introduced by medical jargon, abbreviations, syntactic variations, and context-specific nuances.  The MEDIQA-CORR shared task \citep{mediqa-corr-dataset}, part of the ClinicalNLP 2024 workshop, seeks to address this issue of identifying and correcting (common sense) medical errors found in clinical notes.

The MEDIQA-CORR shared task involves three subtasks: detecting errors in clinical notes, identifying specific error sentences, and correcting those sentences. Our approach involves tackling all three subtasks simultaneously using a single prompt for LLMs invoking chain-of-thought. By doing this, we seek to assess the complex reasoning capabilities of LLMs in the clinical domain.

First, we analyze the dataset and curate a list of the most common types of errors in clinical notes. We then utilize this information to create task specific instructions for the LLM. We leverage contemporary LLMs using in-context learning (ICL) with a few-shot approach. Since publicly accessible LLMs are instruction-tuned models, considering the approach of directing them towards assessing the clinical note based on specific error types maximizes the objective while only utilizing a few training examples.

By employing single-prompt ICL approaches to solve end-to-end tasks, we pave the way forward for building more complex healthcare applications using simple yet intuitive strategies leveraging the capabilities of advanced LLMs.

In addition to the task prompt, an LLM's performance on a particular task is mainly influenced by two factors: training techniques and underlying training data. Consistency, which measures how frequently an LLM produces the same output given the same instructions, can be viewed as another dimension that can affect the quality of an LLM. Leveraging self-consistency can significantly improve the accuracy of an LLM, particularly for complex reasoning tasks \citep{selfconsistency}. In healthcare datasets, where hallucinations in LLMs can occur more frequently due to training on factually unverified data, this could lead to serious problems. While self-consistency is one approach to obtaining more accurate results, LLM ensemble, which has not yet been fully explored, presents a promising opportunity. We validate the results of each LLM by using the output of other LLMs that are trained on different corpora. In our approach, we investigate both self-consistency and ensemble concepts.

The remainder of the paper includes related work, the MEDIQA-CORR task and dataset, our approach, results and conclusion.

\section{Shared Task and Dataset}
The shared task focuses on leveraging LLMs for the following three subtasks:
\begin{itemize}
\setlength\itemsep{0em}
    \item \textbf{Binary Classification:} Detect if the text from a clinical note includes a medical error.
    \item \textbf{Span Identification:} Identify the text span (in the sentence) associated with the error, if an error exists.
    \item \textbf{Natural Language Generation:} Generate the corrected text, if an error exists.
\end{itemize}

\subsection{Subtasks and Metrics}
\subsubsection{Error Detection}
For each clinical text, the prediction is assigned a value of 1 when an error is detected, and 0 if no error is detected. Given the binary nature of this classification task, accuracy serves as the primary metric for performance evaluation.
\subsubsection{Error Span Identification}
Each clinical text comprises sentences associated with unique IDs. The subtask involves predicting the error ID, which is an integer between 0 and the highest sentence ID. If no error is detected, the prediction should be -1. The primary evaluation metric is accuracy, calculated based on all samples, including those with and without errors.
\subsubsection{Correct Sentence Generation}
If a model identifies an error sentence in the previous subtasks, it should also generate a corrected sentence as prediction for this subtask. Here, the full corrected sentence is evaluated against the ground truth sentence for measuring the performance. 
Clinical note generation tasks are challenging to evaluate automatically due to the large number of possible correct answers. Metric ensembles \citep{aggregate} have been shown to outperform individual state-of-the-art measures, such as ROGUE for such tasks. The evaluation metric for this subtask is computed as an unweighted average of the following three scores:
\begin{itemize}
\setlength\itemsep{0em}
    \item \textbf{ROUGE-1F} measures the similarity between system-generated and human-written texts by measuring the overlap of unigrams \citep{rouge}. It uses the F-1 score to assess the quality of the generated sentence.
    \item \textbf{BERTScore} leverages contextual word embeddings obtained from BERT models to assess the similarity between a candidate sentence and a reference sentence \citep{bertscore}. In this context, it signifies the F-1 score of the semantic similarity performed using the DeBERTa XL model \citep{deberta}.
    \item \textbf{BLEURT-20} is a learned metric trained on human ratings that aims to better correlate with human judgments for measuring quality compared to traditional BLEU \citep{bleurt}.
\end{itemize}

\subsection{Dataset}
The train data consists of clinical texts from MS data while the valid and test data contains MS and UW collections. Each entry in the datasets includes a text, its ID and sentences as inputs. Table \ref{tab:dataset} shows the composition of the dataset.

\begin{table}
  \centering
  \begin{tabular}{c c c}
    \hline
    \textbf{Dataset Type} & \textbf{\# Samples} & \textbf{\makecell{\% of Error \\Samples}} \\
    \hline
    MS Training & 2189 & 55.69  \\
    MS Validation & 574 & 55.57 \\
    UW Validation & 160 & 50.00 \\
    MS + UW Test & 925 & 51.35 \\
    \hline
  \end{tabular}
  \caption{MEDIQA-CORR Dataset}
  \label{tab:dataset}
\end{table}

\section{Approach}
We propose to tackle all subtasks concurrently within a single prompt for the following reasons:
\begin{itemize}
\setlength\itemsep{0em}
    \item \textbf{Comprehensive Evaluation:} To enable performance evaluation on a complex task, rather than assessing them on isolated, simpler tasks. This provides a more holistic view of the LLMs' capabilities.
    \item \textbf{Efficiency Optimization:} To minimize inference costs and developmental efforts by eliminating the need for multiple models (or) sequential processing. It streamlines the process, making it more efficient and cost-effective.
    \item \textbf{Ease of Adoption:} To alleviate the adoption burden in practical applications and facilitate seamless upgrades to more advanced LLMs amidst the rapid technological advancements.
\end{itemize}

Publicly accessible LLMs are models that are fine-tuned to follow instructions, with the aim of performing user-defined instructions as accurately as possible. The success of the task then depends on the quality of the instructions provided and the model's ability to follow them effectively. Our approach focuses on refining the instructions for the LLM to facilitate comprehensive learning of all subtasks using ICL. We initiate this process by analyzing error types in the dataset followed by curating the prompt and inferencing with different LLMs.

\subsection{Error Analysis}
\begin{figure}[H]
\caption{Error Types Extracted from Training Data}
\includegraphics[width=8cm]{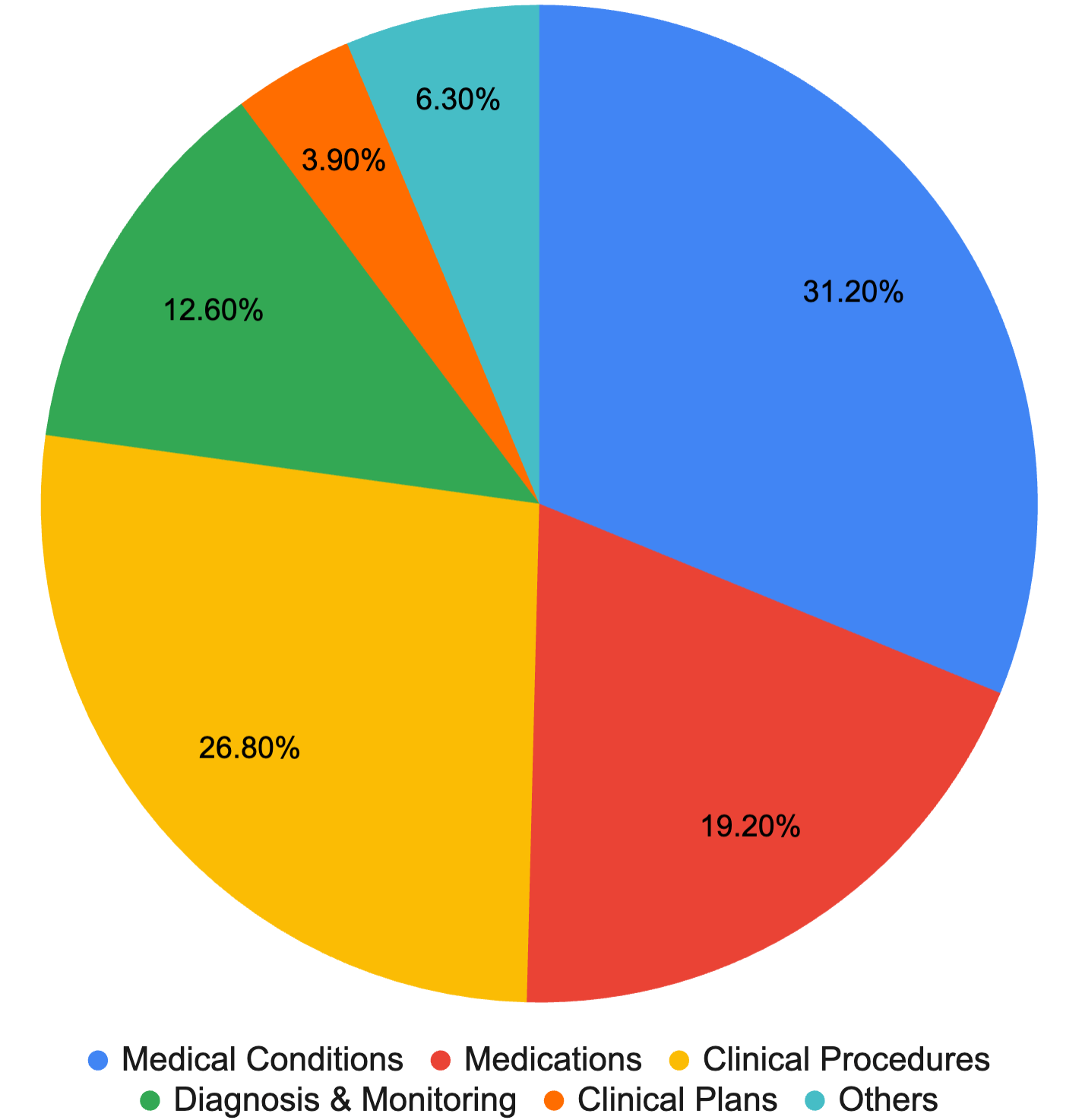}
\label{fig:pie}
\end{figure}

In the MEDIQA-CORR task, the definition of an error is loosely defined and can be interpreted differently by humans or systems without examining the dataset. To address this, we perform error type classification in clinical texts by extracting error sentences and corresponding corrected sentences from the training data. We create an LLM prompting task to broadly categorize the entities modified from the error sentence to the corrected sentence within the clinical domain. We utilize GPT-3.5 for categorizing the errors and cluster these generated free-form categories into a manually defined set. This categorization results in the identification of various error types as depicted in figure \ref{fig:pie}. Finally, we use the well-defined error categories for the rest of the task while handling "Others" category discreetly.

\subsection{Prompt Curation}
When prompted to identify errors in clinical texts without being specific, LLMs may introduce biases from their training data and flag non-critical errors adhering to their own standards of composing clinical notes. To mitigate this, we propose converting an abstract definition of a clinical note error into a concrete and approximate one by analyzing and categorizing the errors. Consequently, we expand the original task from identifying the error to include error classification, which facilitates chain-of-thought for the LLM. We conduct ablation studies to demonstrate the effectiveness of these techniques and present in section \ref{ablation_section}.

To illustrate the task more thoroughly, we incorporate reasoning within the task prompt for the LLM. Through this, we aim to provide more generalizable patterns for detecting and identifying errors. Additionally, it adds explainability to the LLM systems, which is crucial for real-world applications. Finally, we adopt a few-shot approach utilizing random samples from the training data to teach the LLM how to detect, identify, classify errors, provide reasoning and demonstrate the corrected text. We utilize the same samples for few-shot throughout the task as we need to manually generate the additional fields such as error category and reason. The designed task prompt is provided as below. Note that this prompt is a tailored version for demonstration purposes. Finally, errors that fall into the "Others" category are processed as "No Error" as they are unimportant for this task.
\\
\begin{mdframed}
\begin{center} {\bf Prompt Template} \end{center}
In this task, you will be given a clinical report presented as sentences while each sentence is associated with a sentence number. Now, you need to go through the report line by line and identify if there is any error in the sentence. The error can fall into one of the following main categories:\\
1. Medications\\
2. Medical Conditions, Virus or Bacteria\\
3. Reports, Diagnosis and Monitoring\\
4. Clinical Procedures and Treatments\\
5. Clinical Plans and Recommendations\\
6. Medical Devices\\
7. Others including clarity/improper usage of terminology\\
\\
The error can be identified by looking at the entities present in each sentence and check if these entities fall into one of the aforementioned categories and validate if the entire report is correct with this entity. If there is an error, you need to output details as shown in the examples. Use all your medical knowledge and make the right judgements. Here are a few examples for your understanding:
\\ \\
\textit{/* Five random samples from training data with manually curated error category and reason */}\\
\textbf{Example Clinical Report:}\\
0 Mr. <Name> is admitted ..\\
1 He has a surgical ..\\
2 He is also being managed ..\\
\textbf{Output:}\\
\{\\
  "Error Sentence ID": 1,\\
  "Error Category": "Medical Devices",\\
  "Reason": "The device should be .." \\
  "Corrected Sentence": "He has a surgical .."\\
\}\\
...\\
...\\
\textbf{Test Clinical Report:}\\
0 A 45 year old woman ..\\
1 She is experiencing ..\\
2 She had prior examamination ..\\ 
\textbf{Output:}
\end{mdframed}

\subsection{Model Selection}
Using the designed prompt, we utilize the following LLMs for performing the task:
\begin{itemize}
\setlength\itemsep{0em}
    \item \textbf{GPT-3.5:} A model from the OpenAI’s generative pre-trained transformer (GPT) family that can understand as well as generate natural language or code \citep{gpt35}.
    \item \textbf{GPT-4:} Latest model from the GPT family with broader general knowledge and advanced reasoning capabilities \citep{gpt4}. 
    \item \textbf{Claude-3 Opus:} Anthropic’s largest model, released in Feb 2024, which sets new industry benchmarks across a wide range of cognitive tasks and outperforms its peers on most of the common evaluation benchmarks for AI systems \citep{opus}.
\end{itemize}

Due to its affordability, speed, and reliability, GPT-3.5 is an excellent choice for experimentation. As a result, we employed GPT-3.5 for error analysis and prompt design, reserving the more advanced GPT-4 and Claude-3 Opus models for the final test runs.

\subsection{Enhancing Robustness}
Due to the potential limitations such as hallucinations and inconsistent results, which can affect the quality of the LLMs, we investigate two concepts to improve performance on the subtasks: self-consistency and ensemble. The Claude-3 Opus model has slower speed, higher cost, and stricter token limits compared to GPT-4. Therefore, we utilize GPT-4 for self-consistency by generating four outputs per test sample and aggregate them, while only generating one output per test sample for Claude-3 Opus.

\begin{figure*}[!ht]
\caption{LLM Ensemble Approach for MEDIQA-CORR task}
\includegraphics[width=\textwidth]{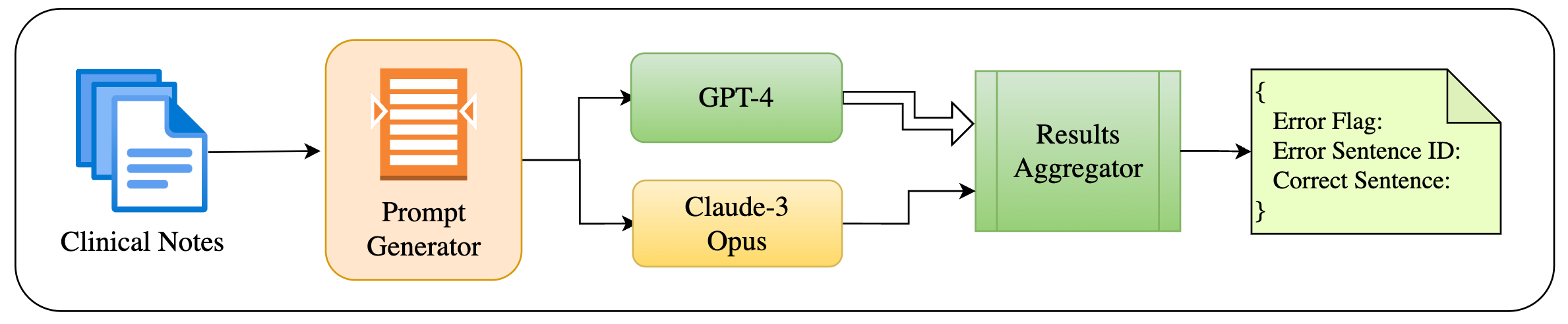}
\label{fig:pipeline}
\end{figure*}

To enhance the quality of predictions, we combine the results from both models to predict the outputs for all three subtasks. Figure \ref{fig:pipeline} provides a visual representation of the overall process. The results aggregator module validates and combines the outputs i.e. predicted error flag, predicted error sentence ID and corrected sentence, from GPT-4 and Claude-3 Opus models to generate the final error flag, error sentence ID and corrected sentence.

\section{Ablation Study} \label{ablation_section}
We begin by presenting the results obtained from incorporating error categorization into the final prompt, which demonstrates an improvement in performance on both error detection and text span identification tasks. In order to make the comparison, we utilize GPT-3.5 with prompts including and excluding error categorization. Additionally, we assess the performance of GPT-4 to ascertain the extent to which it surpasses GPT-3.5 for the finalized prompt. The results obtained using the combined MS and UW validation sets (during the development phase) are presented in Table \ref{tab:ablation}.

\begin{table}
  \centering
  \resizebox{\columnwidth}{!}{%
  \begin{tabular}{c c c c}
    \hline
    \textbf{Model} & \textbf{Prompt} & \textbf{\makecell{Task-1\\ Accuracy}} & \textbf{\makecell{Task-2\\ Accuracy}} \\
    \hline
    GPT-3.5 & \makecell{No error\\ categories} & 48.75\% & 22.5\%  \\
    \hline
    GPT-3.5 & \makecell{Error\\ categories} & 58.44\% & 38.55\% \\
    \hline
    GPT-4 & \makecell{Error\\ categories} & 63.07\% & 58.17\% \\
    \hline
  \end{tabular}
  }
  \caption{Performance improvement with error categorization in prompt using GPT-3.5 and assessing GPT-4 performance}
  \label{tab:ablation}
\end{table}

The results indicate that by integrating error categorization which initiates an intermediate chain-of-thought, results in a significant performance boost of nearly 10\% and 16\% for Task-1 and Task-2, respectively. Additionally, GPT-4 outperforms GPT-3.5, confirming its enhanced reasoning capabilities. These advancements make GPT-4 a preferred choice for final test runs.

\section{Results} \label{results_section}
We present the results of individual models first followed by incorporating self-consistency and ensembles on the test data. The performance of GPT-4 and Claude-3 Opus using the final prompt on all three subtasks is presented in Table \ref{tab:result1}.

\begin{table}
  \centering
  \resizebox{\columnwidth}{!}{%
  \begin{tabular}{c c c c}
    \hline
    \textbf{Model} & \textbf{\makecell{Task-1\\ Accuracy}} & \textbf{\makecell{Task-2\\ Accuracy}} & \textbf{\makecell{Task-3 Agg-\\ regate Score}} \\
    \hline
    GPT-4 & 62.05\% & 56.43\% & 0.6172  \\
    \hline
    \makecell{Claude-3\\ Opus} & 62.59\% & 58.48\% & 0.6669 \\
    \hline
  \end{tabular}
  }
  \caption{Comparison of GPT-4 and Claude-3 Results}
  \label{tab:result1}
\end{table}

Claude-3 Opus surpasses GPT-4 in error detection and significantly in error sentence identification. However, GPT-4 tends to be more verbose during error correction, leading to lower scores in metrics such as ROUGE, BERT, and BLEURT. Although Claude-3 Opus exhibits superior performance, its daily token limit, slower inference, and shorter test phase duration hinder its usability for self-consistency. Therefore, we employ GPT-4 to demonstrate the performance enhancement using self-consistency in large language models (LLMs). Additionally, we ensemble the self-consistent GPT-4 with Claude-3 Opus to showcase further improvement. Table \ref{tab:result2} presents the results for all subtasks using the aforementioned methods:

\begin{table}
  \centering
  \resizebox{\columnwidth}{!}{%
  \begin{tabular}{c c c c}
    \hline
    \textbf{Model} & \textbf{\makecell{Task-1\\ Accuracy}} & \textbf{\makecell{Task-2\\ Accuracy}} & \textbf{\makecell{Task-3 Agg-\\ regate Score}} \\
    \hline
    \makecell{GPT-4 with\\ consistency\\ (Majority=3/4)} & 62.91\% & 59.45\% & 0.6390  \\
    \hline
    \makecell{GPT-4 +\\ Claude-3 Opus\\ (Majority=4/4)} & 62.16\% & 60.86\% & \textbf{0.7865}  \\
    \hline
    \makecell{GPT-4 +\\ Claude-3 Opus\\ (Majority=3/4)} & \textbf{63.78\%} & \textbf{62.48\%} & 0.7492  \\
    \hline
  \end{tabular}
  }
  \caption{Self-consistent GPT-4 and its ensemble with Claude-3 Opus Results}
  \label{tab:result2}
\end{table}

The majority ratio x/y for GPT-4 results is a measure of how often the model produces the same result for a given input. For example, a majority ratio of 3/4 means that at least three out of the four results should be the same to qualify the predicted error sentence ID as correct. In the ensemble approach, a prediction is considered correct if the self-consistent GPT-4 result matches the Claude-3 Opus result. Otherwise, the prediction is considered "no error". To select the best corrected sentence from the ensemble, the ROGUE score is used to estimate the distance between each corrected sentence and the error sentence. The sentence with the highest ROGUE score is used for the evaluation because LLMs tend to generate verbose corrected sentences which decreases the aggregate scores. In our experiments, using a majority ratio of 3/4 for GPT-4 in the ensemble resulted in the best Task-1 and Task-2 performances. Using a majority ratio of 4/4 resulted in the best Task-3 aggregate score. Our best subtask-3 score is ranked 2nd among all participants in the competition, and our best subtask-2 performance is among the top 3 according to the reported scores \citep{mediqa-corr-task}.

\section{Related Work}
In recent years, there has been a surge of research exploring the potential of prompt engineering techniques with large language models (LLMs) in healthcare. These techniques have shown promising results in various healthcare tasks, often achieving state-of-the-art performances \citep{surveyhealthcare}, \citep{surveyhealthcare2}. One notable study, MedPrompt, highlighted several research directions demonstrating the power of prompt exploration for LLMs \citep{generalist}. LLMs exhibited impressive knowledge and reasoning abilities, tackling various tasks effectively. These advancements showcase the potential of LLMs in healthcare, offering new opportunities for leveraging language models to address healthcare challenges.

Evaluating common sense reasoning is essential for computer systems, as it impacts language comprehension, communication reliability, and general task performance. SemEval-2020 ComVE aims to address general common sense questions and seeks logical justification for correct responses, assessing reasoning abilities. Pretrained language models naturally acquire common sense through training on vast word tokens obtained from the web \citep{semeval}. MEDIQA-CORR, specifically tailored to identify and correct errors in clinical notes, offers a valuable resource for evaluating pretrained LLMs in medical common sense reasoning. Inspired by prompt-based explorations, our research also focuses on utilizing pretrained LLMs to assess reasoning capabilities in medical common sense scenarios.

\section{Conclusion}
Our research demonstrates that incorporating error categorization into the prompt enhances the performance of large language models (LLMs) in detecting, identifying, and classifying clinical note errors. By initiating an intermediate chain-of-thought, this approach facilitates better reasoning and aids the LLM in providing more accurate and explainable results. Furthermore, our findings suggest that self-consistency and ensembles can further enhance the robustness and performance of LLMs on these tasks. These advancements pave the way for the development of more reliable and interpretable AI systems for clinical documentation analysis, ultimately contributing to improved healthcare outcomes.

\section{Limitations and Risks}
While promising, our approach has limitations. LLMs trained on general data may lack specific medical knowledge, potentially leading to misinterpretations and inaccurate corrections. Despite mitigation efforts, the risk of hallucinations and inconsistencies in LLM outputs remains a concern. Additionally, the effectiveness of our approach relies heavily on prompt engineering, which requires expertise and may not be easily generalizable.

The black box nature of LLMs also presents challenges in terms of explainability and building trust in medical contexts. To mitigate these limitations, continuous improvements of training data, more robust evaluation metrics, and human oversight are crucial. Further research is needed to explore the full potential and limitations of LLMs in healthcare, ensuring their safe and responsible application for improved patient care.

\section{Ethical Considerations}
The use of LLMs for medical error detection and correction raises significant ethical concerns. Potential biases in training data and algorithms must be carefully mitigated to prevent propagating existing healthcare disparities and ensure fairness. Transparency in how prompts are designed and how the LLM reaches its decisions is vital for building trust and ensuring accountability. Additionally, robust data security and de-identification practices are paramount for protecting sensitive patient information.

It is essential to remember that LLMs should serve as tools to augment the expertise of healthcare professionals, not replace them. Clear lines of responsibility, ongoing human oversight, and continuous research and collaboration are necessary to address these ethical challenges. This will ensure the responsible use of LLMs and their positive contribution to improved healthcare outcomes.

\bibliography{custom}

\end{document}